\def\year{2020}\relax
\documentclass[letterpaper]{article} 
\pdfoutput=1 
\usepackage{aaai20}  
\usepackage{times}  
\usepackage{helvet} 
\usepackage{courier}  
\usepackage[hyphens]{url}  
\usepackage{graphicx} 
\usepackage{graphicx}  
\title{Is it a Fruit, an Apple or a Granny Smith? Predicting the Basic Level in a Concept Hierarchy}
\iftrue
\author{Laura Hollink, Aysenur Bilgin, Jacco van Ossenbruggen\\  
Centrum Wiskunde \& Informatica\\
Science Park 123, 1089XG Amsterdam, The Netherlands\\
laura.hollink@cwi.nl, aysenur.bilgin@cwi.nl, jacco.van.ossenbruggen@cwi.nl 
}
\fi
 
\begin{document}

\maketitle

\begin{abstract}




The ``basic level'', according to experiments in cognitive psychology, is the level of abstraction in a hierarchy of concepts at which humans perform tasks quicker and with greater accuracy than at other levels. We argue that applications that use concept hierarchies -- such as knowledge graphs, ontologies or taxonomies -- could significantly improve their user interfaces if they `knew' which concepts are the basic level concepts. This paper examines to what extent the basic level can be learned from data. We test the utility of three types of concept features, that were inspired by the basic level theory: lexical features, structural features and frequency features. We evaluate our approach on WordNet, and create a training set of manually labelled examples that includes concepts from different domains. Our findings include that the basic level concepts can be accurately identified within one domain. Concepts that are difficult to label for humans are also harder to classify automatically. Our experiments provide insight into how classification performance across domains could be improved, which is necessary for identification of basic level concepts on a larger scale. 
\end{abstract}
\section{Introduction}
One of the ongoing challenges in Artificial Intelligence is to explicitly describe the world in ways that machines can process. This has resulted in taxonomies, thesauri, ontologies and more recently knowledge graphs. 
While these various knowledge organization systems (KOSs) may use different formal languages, they all share similar underlying data representations. 
They typically contain instances 
and classes, or concepts, and they use subsumption hierarchies to organize concepts from specific to generic. 



In this paper, we aim to enrich concept hierarchies by predicting which level of abstraction is the `basic level.' This is a notion from the seminal paper by \citeauthor{Rosch1976a} (\citeyear{Rosch1976a}) in which they present the theory of ``basic level categories.''\footnote{Note that vocabulary varies per research community and throughout time. Rosch's ``categories'' would be called ``classes'' or ``concepts'' in recent Knowledge Representation literature.} 
The core idea is that in a hierarchy of concepts 
there is one level of abstraction that has special significance for humans. At this level, humans perform tasks quicker and with greater accuracy than at superordinate or subordinate levels. In a hierarchy of edible fruits, this so called `basic level' is at the level of \textit{apple} and not at the levels of \textit{granny smith} or \textit{fruit}; in a hierarchy of tools it is at the level of \textit{hammer} rather than at the levels of \textit{tool} or \textit{maul}; and in a hierarchy of clothing it is at the level of \textit{pants}. In a series of experiments, Rosch demonstrated that humans consistently display `basic level effects' -- such as quicker and more accurate responses -- across a large variety of tasks. 

In contrast, in current knowledge graphs and other KOSs each level in the hierarchy is treated equally. To illustrate why this may be problematic, consider the following example. Using a taxonomy of fruits combined with the fact that an image displays a golden delicious, we can infer new facts: that it displays an apple and that it displays a fruit. However, that doesn't tell us which is the best answer to the question ``What is depicted?''-- a golden delicious, an apple or a fruit? In cases where the concept hierarchy is deep, there might be dozens of concepts to choose from, all equally logically correct descriptions of the image. KOSs generally have no mechanism for giving priority to one level of abstraction over another.

We argue that applications that use knowledge graphs could significantly improve their user interfaces if they were able to predict the basic level concepts. In other words, if they were able to predict for which concepts in the graph users can be expected to display basic level effects. In the example above, we illustrated how computer vision systems could be designed to describe the objects they detect at the basic level rather than at subordinate or superordinate levels, so that users can react as quickly as possible. Another example is an online store, that could cluster products at the basic level to give users a quick overview of what is sold, rather than choosing more specific or more general clusters. Automatic summarization systems could describe the contents of an article at the basic level, etc. It is important to note that we do not argue that the basic level should \textit{always} be the preferred level of abstraction. For example, in indexing as it is done in digital libraries, it is often advisable to select concepts as specific as possible. Also, in application-to-application situations where there is no interaction with a human user, basic level effects are obviously irrelevant.

Motivated by these example scenarios, our goal is to predict which concepts in a given concept hierarchy are at the basic level. We do this in a data-driven manner, in contrast to the laboratory experiments with human users that have been conducted in cognitive psychology. 


We train a classifier using three types of features. Firstly, we elicit lexical features like word-length and the number of senses of a word, since it has commonly been claimed that basic level concepts are 
denoted by shorter and more polysemous words \cite{Murphy1982,Tanaka1991,Green2006}. 
Secondly, we extract structural features, such as the number of subordinates of a concept and the length of its description. This is motivated by a definition of the basic level being ``the level at which categories carry the most information'' \cite{Rosch1976a}. Finally, we obtain features related to the frequency of use of a word, since basic level concepts are thought to be used often by humans. 

To test our approach, we apply it to the concept hierarchy of WordNet, a widely used lexical resource, and classify WordNet concepts as basic level or not-basic level. For training and testing, we create a gold standard of 518 manually labelled concepts spread over three domains. Frequency features are extracted from Google Ngram data. Lexical and structural features are extracted from the concept hierarchy itself, i.e. from WordNet. In a series of experiments, we aim to answer three research questions: 
1) to what extent can we predict basic level concepts within and across domains, 2) how can we predict the basic level in new, previously unseen domains, and 3) how does machine classification compare to human classification, i.e. what is the meaning of disagreement between human annotators?
We believe the answer to these two questions will bring us one step closer to the overall aim of being able to predict the basic level on a large scale in all kinds of concept hierarchies, helping applications built on top of them can interact with users more effectively.



\section{Related Work}\label{sec:relatedwork}


\citeauthor{Rosch1976a} (\citeyear{Rosch1976a}) demonstrated basic level effects across a large variety of tasks. For example, they found that people, when asked to verify if an object belonged to a category, reacted faster when it was a basic level category (``Is this a chair'' is answered quicker than ``Is this furniture?''); when asked to name a pictured object, people chose names of basic level concepts (They said ``It is an apple'' rather than ``It is a golden delicious'');  and when asked to write down properties of a concept, people came up with longer lists if the concept was at the basic level (many additional properties were named for ``car'' compared to the properties of it's superordinate ``vehicle'', while few additional properties were mentioned for ``sports car''). In the present paper, we aim to derive `basic levelness' in a data driven manner, rather than by performing psychological experiments, to allow for enrichment of concept hierarchies at a large scale.

Rosch's initial experiments were done on a relatively small set of nine hierarchies of ten concepts each. She chose common, tangible concepts, such as fruits, furniture and musical instruments. Later, basic level effects were also demonstrated in other types of concepts, such as events \cite{Rifkin1985}, geographical features \cite{Mark1999}, sounds \cite{Lemaitre2013} and categories with artificially created names \cite{Murphy1982}. 
These results show that basic level effects exist on a much wider scale than Rosch's relatively clear\-cut examples, strengthening our claim that there is a need to automatically derive the basic level in large concept hierarchies. 



The basic level is relatively universal since it is tightly coupled with universal physiological features such as what humans can perceive and what movements they are capable of \cite{lakoff2008women}. 
That being said, it should be acknowledged that there are also individual factors that affect to what extent basic level effects occur. One of those factors is expertise. Domain experts may process subordinate levels with equal performance in the context of their domain of expertise \cite{Johnson1997,Tanaka1991}. Similarly, the familiarity of a person with an object plays a role, where familiarity increases basic level effects \cite{Smith1967}. Finally, the proto\-typicality of an object is a factor; if someone perceives an object as a prototypical example of its class, basic level effects may increase \cite{Rosch1976}. These individual factors are outside the scope of the present paper, where we focus on the universal nature of basic level effects. 

The idea of a `basic level' has been used in various applications that use conceptual hierarchies.
In the context of the semantic web, for example, it has been used in ontology creation \cite{Uschold1995,Hoekstra2007}, automatic ontology generation \cite{Golder,clerkin2001ontology,Cai2016}, ontology matching \cite{Green2006} and entity summarization \cite{PeroniMd08}. 

\citeauthor{ordonez2013large} (\citeyear{ordonez2013large}) stress the importance of the basic level in computer vision. They propose a mapping between basic level concepts and the vocabulary of concept names that is used by existing visual recognition systems. \citeauthor{mathews2015choosing} (\citeyear{mathews2015choosing}) use collaborative tags to predict basic level names of recognized objects in an automatic image captioning task. 

For all these applications there is a need to identify which concepts are at the basic level. In the papers mentioned above this was done either manually \cite{Uschold1995,Hoekstra2007}, using heuristics \cite{PeroniMd08,Green2006}, by looking at the frequency and order of occurrence of user generated tags \cite{Golder,ordonez2013large,mathews2015choosing}, or using a measure called category utility \cite{Cai2016,clerkin2001ontology}. 

The category utility \cite{corter1992explaining} of a concept $c$ is a measure of how well the knowledge that item $i$ is a member of $c$ increases the ability to predict features of $i$. For example, knowledge that $i$ is a bird allows one to predict that $i$ can fly, has wings, lays eggs, etc. \citeauthor{Belohlavek2013} (\citeyear{Belohlavek2013}) compared the category utility measure for basic level prediction to two similar measures that were proposed earlier, such as cue validity \cite{Rosch1976a} and Jones' category-feature collocation measure (\citeyear{jones1983identifying}), and found that they lead to similar predictions.

In contrast to category utility, cue validity and the category-feature collocation measure, our approach does not rely on the availability of explicit information about all features of a concept. In our experience, features such ``can fly'' and ``has wings'' are seldom encoded in a concept hierarchy. Our approach builds on the idea of using tag frequency by including the frequency of occurrence of a concept in a natural language text corpus as a feature. Finally, our approach is inspired by some of the heuristics proposed before, e.g. with respect to the use of depth in the hierarchy \cite{PeroniMd08} and lexical properties \cite{Green2006}.

\section{A Method for Basic Level Prediction}
This section presents basic level prediction as a binary classification task. It describes the features that are used as input to the classifier, and discusses how concepts are manually labelled for training and testing purposes.

\subsection*{Task}
\label{sec:task}
We cast the task of basic-level identification as a binary classification problem: a concept in a hierarchy either is or is not at the basic level. 
In future work, we intend to look into a multi-class classification task, distinguishing basic level, more specific and more generic concepts. In many cases, the identification that a concept is more specific/generic is trivial, as it follows directly from the position in the hierarchy relative to the basic level concept. However, in cases where a branch does not include a basic level concept, the task is nontrivial and a distinction between more specific and more generic concepts may be valuable to 
applications. 

\begin{figure}
  \centering
  \includegraphics[width=0.7\columnwidth]{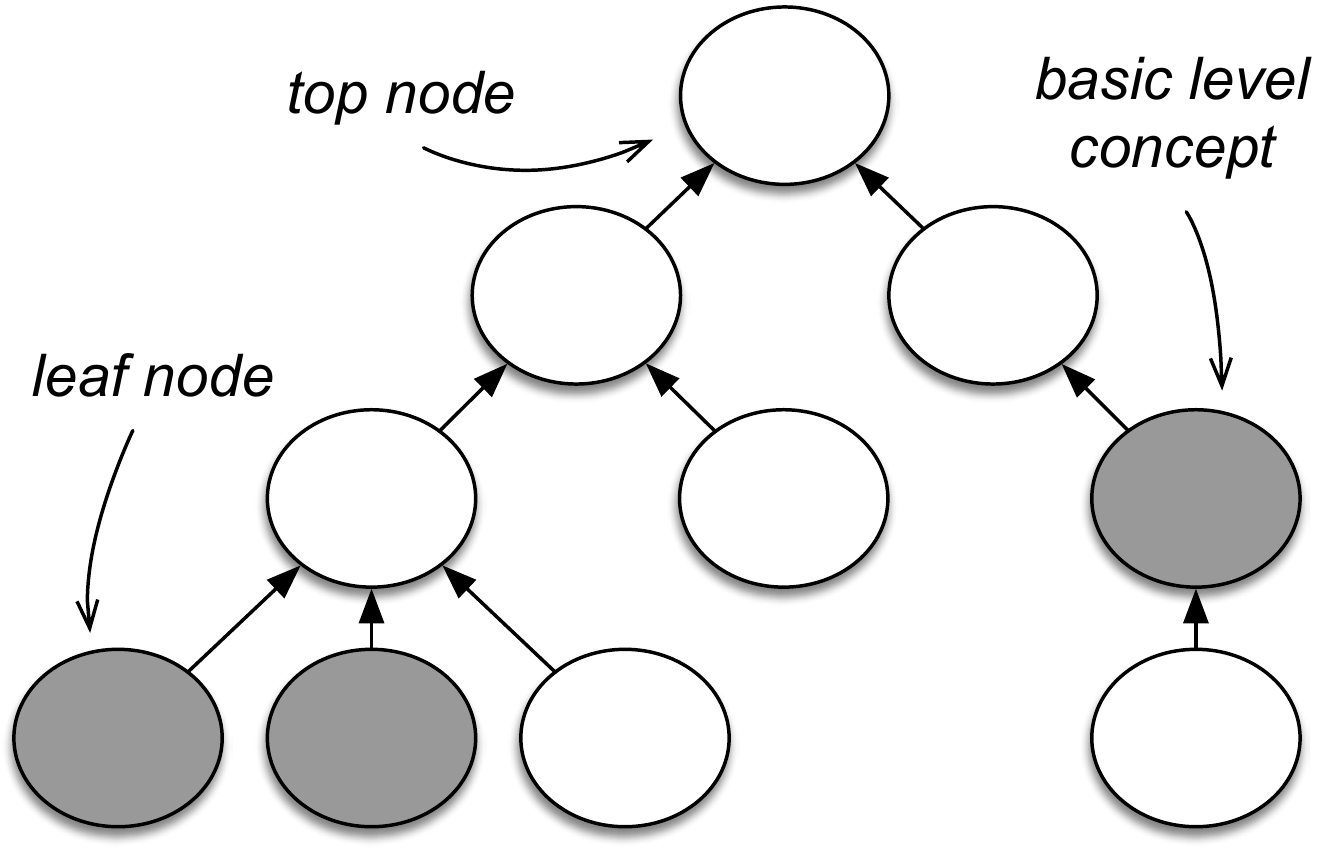}
  \caption{Example hierarchy, basic level concepts in grey.}
  \label{fig:hierarchy}
\end{figure}

Figure \ref{fig:hierarchy} shows an example hierarchy in which the basic level concepts have been identified. The figure illustrates that the basic level concepts can be at different levels in the hierarchy for different branches. Each branch from the top node to a leave node has at most one basic level concept.

\subsection*{Extracting Three Types of Features}
As input to the classifier, we extract structural, lexical and frequency features. 
We use the conceptual hierarchy to extract structural features about a concept. \citeauthor{Rosch1976a} (\citeyear{Rosch1976a}) showed that at the basic level people tend to give longer descriptions when asked to describe a concept and were able to list most new attributes at this level. They concluded that the basic level is ``the level at which categories carry the most information.'' Based on this, we hypothesize that the amount of explicit information about a concept in a KOS can be used as a signal for basic level prediction. Accordingly, we collect the number of relations that the concept has in the knowledge graph or other KOS. Depending on the conceptual hierarchy, these can be is-a relations such as \verb|rdfs:subClassof|, \verb|skos:broader| or \verb|wordnet:hyponym|. If other types of relations are present in the KOS, such as part-of relations, we count these as well. If the conceptual hierarchy contains natural language descriptions of the concepts, we include the length of the description as a feature. We also store the depth of the concept in the hierarchy, measured as the shortest path of is-a relations from the concept to the top node. 

The use of lexical features is motivated by the fact that the basic level can be recognized in natural language. Many have claimed that basic level concepts are generally denoted by shorter and more polysemous words \cite{Murphy1982,Tanaka1991,Green2006}. 
They are also the first words acquired by children \cite{brown1958shall}.  
The extraction of lexical features requires a mapping from concepts to words. In knowledge representation languages commonly used for KOSs, this mapping is for example given by the \verb|rdfs:label| or \verb|skos:preferredLabel| relations. We extract the following lexical features for each concept: the length of the word(s) (measured in characters), the number of senses of the word(s) (i.e. polysemy) and the number of synonyms of the word.

Finally, we include the frequency of occurrence of a concept as a feature. 
Rosch's initial experiments \cite{Rosch1976a} demonstrated that people often choose basic level concepts when asked to describe an object, rather than subordinate or superordinate concepts. Therefore, we hypothesize that the frequency of occurrence of a word in a corpus of natural language text is a signal for basic level prediction.

Both lexical and frequency features are based on words. In many KOSs, one concept can be denoted by multiple words. If this is the case, we need to aggregate multiple word-level feature values into one concept-level feature value. We use mean, minimum or maximum values for this purpose. 
For example, when a concept is denoted by two synonyms -- let's say \textit{piano} and \textit{pianoforte} -- we could use the mean word-length (7.5 characters) but also the minimum word-length (5 characters) as a concept-level feature. 
What we call a ``word'' here, may actually be a multi-word phrase. For example, in a hierarchy of musical instruments, the largest of the violin-type instruments is denoted by the word \textit{contrabass} as well as by the ``word'' \textit{double bass}. We treat multi-word phrases the same as single words.

As we rely on manually labelled examples for training and testing, our data set is relatively small. Therefore, we aim to also keep the number of features relatively low. In order to handle (multi)collinearity, we choose to remove highly correlated features 
(in this study, Spearman's $\rho > 0.75$). High correlations are expected between the various aggregation methods of word-based features (e.g. between the mean word-length and minimum word-length). When two features have a strong correlation, we keep the feature that discriminates most between basic level and not-basic level concepts and discard the other. 

\subsection*{Creating Manual Concept Labels for Training and Testing}
We ask human annotators to manually label concepts as being basic level or not. For this purpose, an annotation protocol was provided that includes a short description of what the basic level is, as well as the results from Rosch's initial experiments. For each concept, we provide the annotators with the synonyms that denote the concept, the position in the hierarchy, and a natural language description of the concept. The protocol lists additional sources that the annotator may consult: Wikipedia, Google web search and/or Google image search, all with a standardized query consisting of a word that denotes the concept and optionally a word of a superordinate concept. Finally, the protocol provides some hints that may help the annotator to decide on the correct label in case of doubt. For example, ``at the basic level an item can often easily be identified even when seen from afar'', ``at the basic level there is often a recognisable movement associated with the concept,'' and ``at the basic level images of the item often all look alike'' 
Cohen's $\kappa$ as well as Krippendorf's $\alpha$ were used to determine the inter-rater agreement. Where $\alpha$ is equal to $\kappa$ (when rounded to 2 decimal points), we report $\kappa$ only. 

With this manual labelling method, we deviate from previous work. While Rosch and other cognitive psychologists \textit{measured} basic level effects in laboratory experiments, we simply ask annotators whether they \textit{think} a concept is at the basic level. 
This pragmatic choice allows us to create a larger training and test set. We work under the assumption that with a good knowledge of the basic level literature, as well as an extensive annotation protocol, our human annotators can produce labels that are close to what laboratory experiments would produce. High inter-rater agreement scores (reported in section \ref{sec:trainingtestset}) support this. In future work, however, we intend to test this assumption by measuring basic level effects on a large scale in a crowd-sourcing environment.


\section{Experiments and Evaluation}

\begin{figure}[t]
  \centering
  \includegraphics[width=\columnwidth]{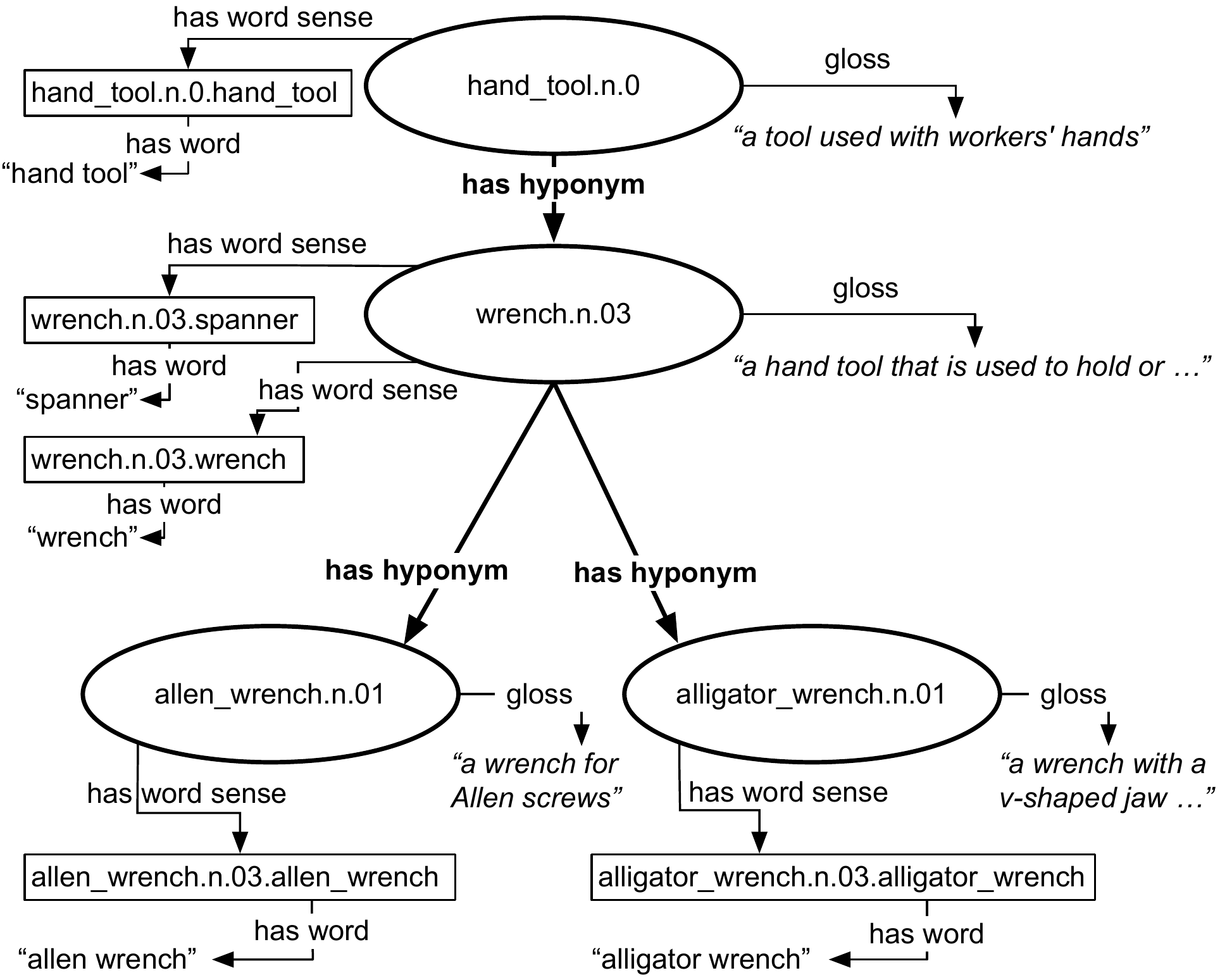}
  \caption{(Partial) data structure of WordNet.}
  \label{fig:wordnet}
\end{figure}

We apply our basic level prediction method to part of WordNet, a lexical database of English 
\cite{miller1995wordnet}. It contains 155k words, organized into 117k synsets\footnote{\url{https://wordnet.princeton.edu/documentation/wnstats7wn}}. A synset can be seen as a concept in a KOS. It is a set of words that denote the same concept, i.e. a set of synonyms. Synsets are connected to each other through the \verb|hyponym| relation, which is an is-a relation. Figure \ref{fig:wordnet} shows an example of four synsets in a hyponym hierarchy. Each synset has a natural language description called a \verb|gloss|. Synsets are connected to (one or more) words via word senses. 

WordNet presents a particularly interesting test ground considering (1) the depth of its hierarchy, making the identification of the basic level challenging and valuable, (2) its wide scope that includes all concepts that Rosch used in her original experiments, and (3) its widespread use. 

The task then can be formulated to be classification of synsets in WordNet as basic level or not. We perform experiments on a subset of WordNet consisting of 518 manually labelled noun synsets. 

The dual role of WordNet as a lexical database and a conceptual hierarchy allows us to extract both lexical features and structural features from it. For frequency features, we use Google Ngram data \footnote{https://books.google.com/ngrams}.

\subsection*{Extracting Structural Features}
We extract the following structural features from WordNet:


\begin{itemize}
    \item \textbf{nr\_of\_hyponyms} Hyponym, the main is-a relation in WordNet, is a transitive relation. We count the number of synsets in the complete hyponym-tree under the synset
    \item \textbf{nr\_of\_direct\_hypernyms} 
    Hypernym is the inverse relation of hyponym. As WordNet allows for multiple classification, some synsets have multiple hypernyms. We count We count the number of hypernyms directly above the synset.
    \item \textbf{nr\_of\_partOfs} The number of holonym plus meronym relations of the synset.
    \item \textbf{depth\_in\_hierarchy} The number of hyponyms in the shortest path from the synset to WordNet's root noun 
    \item \textbf{gloss\_length} The nr. of characters in the synset gloss. 
\end{itemize}{}

During the feature selection step, one structural feature was pruned, namely the number of direct hyponyms, because of a strong correlation to nr\_of\_hyponyms ($\rho = 0.80$).

%

\subsection*{Extracting Lexical Features}
We extract word-length, polysemy and synonymy from WordNet. Word-length and polysemy are both word-based features. When a synset contains multiple words, the word-based features need to be aggregated into synset-based features. For word-length, we tested the mean and the minimum length, based on the expectation that basic level concepts are denoted by shorter words. For polysemy, we tested the mean and the maximum number of synsets in which the words occur, since at the basic basic level words are said to be the most polysemous. After pruning strongly correlated features ($\rho = 0.87$ between the two word-length features and $0.89$ between the two polysemy features), minimum word length and maximum polysemy were kept as features. This results in the following set of lexical features:

\begin{itemize}
    \item \textbf{word\_length\_min} The number of characters of the shortest word in the synset.
    \item \textbf{polysemy\_max} The number of synsets in which the most polysemous word of the synset appears.
    \item \textbf{nr\_of\_synonyms} The number of words in the synset.
\end{itemize}

\subsection*{Extracting Frequency Features from Google Ngrams}
Google Ngrams provides data about how often a word -- or a multi-word phrase, an ``n-gram'' -- appears in the Google Books corpus. This is expressed as the number of times a given ngram appears in a given year, divided by the total number of ngrams in that year. In the present study, the Google Books corpus `English 2012' was used, which comprises of 4.5M books in the English language published between 1800 and 2008 \cite{lin2012syntactic}. We collect the ngram score for each word in the synset. 
 
We initially use two types of aggregations from word-based values to concept-based values, namely the mean and the maximum. In addition, we collect data from two corpora: the entire Google Ngram corpus and the sub-corpus of books published in the most recent year available, which is 2008. This results in four variants of the frequency feature. As expected, the four variants correlate strongly ($\rho$ between 0.97 and 0.99). After pruning, one frequency feature remains:

\begin{itemize}
    \item \textbf{G.Ngrams\_score\_2008\_max} the ngram score of the most frequent word in the synset in the most recent Google Books Ngram data.
\end{itemize}    

\subsection*{Training and Test-Set}\label{sec:trainingtestset}
The training and test set consists of synsets from three different parts of WordNet: the complete hierarchies under the synsets \verb|hand_tool.n.01|, \verb|edible_fruit.n.01| and \verb|musical_instrument.n.01|. In this paper, we will refer to these three hierarchies as ``domains.'' They correspond to three of the six non-biological hierarchies that Rosch reported in her seminal paper on basic level effects \cite{Rosch1976a}. The WordNet domains used in the present paper, however, are larger than Rosch's experimental data; they consist of 150+ concepts per domain, whereas Rosch's hierarchies consisted of 10 concepts each. All synsets were manually labelled by two annotators (two of the authors). 

Table \ref{tab:gold} lists the properties of the training and test set including the inter-rater agreement  ($\kappa$). The agreement is substantial, 
with some variation between the domains: $\kappa$ is higher for tools and fruit than for the musical domain. This data set is available from [URL removed for blind review].


\begin{table}
  \caption{Properties of the training and test set: the number of synsets, the maximum depth of the hierarchy counted from (and including) the top synset of the domain, the number of basic level concepts, and the inter-rater agreement ($\kappa$).}\smallskip
  \centering
  \smallskip\begin{tabular}{l|rrrr}
    \hline
    Top-synset & Size & Max. &  \#basic & $\kappa$ \\
    of domain &\#syns.& depth &level& \\
    \hline
    Hand tool      & 157 & 6 & 30 & 0.73\\
    Edible fruit   & 197 & 5 & 77 & 0.78\\
    Musical Instr. & 164 & 7 & 54 & 0.64\\
    \hline
    All            & 518 & 7 & 161 & 0.73\\
    \hline
  \end{tabular}
  \label{tab:gold}
\end{table}

In experiments 1 and 2 below, we will use the subset of the synsets for which both annotators agreed (453 out of 518 synsets). The rationale is that this is a high-quality gold standard, allowing for more reliable insights. In experiment 3, we will evaluate to what extent this choice affects the results, and explore the cases where annotators disagreed.


\subsection{Experiments}
In a series of experiments, 
we measure performance on the binary classification task described in Section \ref{sec:task}. 
Performance is measured with balanced accuracy, precision, recall, F1 and Cohen's $\kappa$. In experiments 1 and 3, we report the distribution of performance scores using a 10-fold cross-validation setup. Learning is performed with Random Forests, using an off-the-shelf toolkit\footnote{The CARET Library in R \url{http://topepo.github.io/caret/index.html}}. We use the SMOTE algorithm implemented the toolkit to deal with class imbalance. 
Other classification algorithms -- Support Vector Machines, Linear Discriminant Analyses, K-Nearest Neighbors and Decision Trees -- were tried but the Random Forest performed the best in almost all cases. For brevity, these experiments have been left out. 

\subsubsection*{Experiment 1: Prediction Within and Across Domains}
In this experiment, we compare 10-fold cross validation results of a global model that is trained and tested on the entire data-set against those of local models, trained and tested on a single domain (tools, fruit, or music)

\subsubsection*{Experiment 2: Prediction in a New Domain}
We examine to what extent a trained model can be used to identify the basic level in a new, previously unseen domain. To simulate this situation, we train on two domains, and test on a third. We train on tools+fruit and test on music; we train on fruit+music and test on tools; and we train on tools+music and test on fruit.

\subsubsection*{Experiment 3: Disagreement Among Human Annotators} 
We compare prediction performance when training and testing is done on the manual labels given by annotator 1, those given by annotator 2, and on the subset of manual labels that they both agree on. 



\section{Results}

\subsubsection{Experiment 1: Prediction Within and Across Domains}

\begin{figure*}
  \centering
  \includegraphics[width=\linewidth]{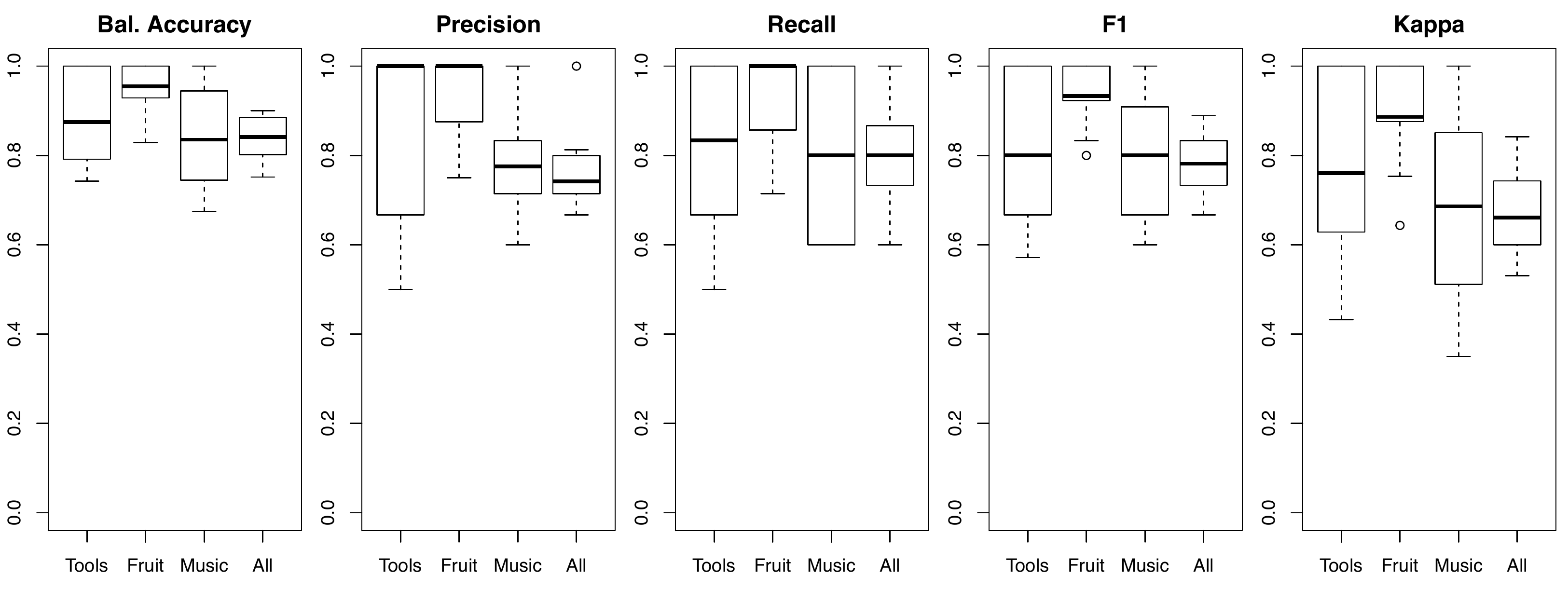}
  \caption{Prediction performance measures on single domains (tools, fruit, music) and on the entire data set (all).} \label{fig:exp1}
\end{figure*}

Figure \ref{fig:exp1} shows prediction performance measures of local models trained and tested on a single domain (tools, fruit, or music) as well as performance of a global model on the entire data set (all). The box-plots summarize the 10-fold cross-validation results. There is some variation between the domains. Performance in the Fruit domain is best on all measures (median bal. accuracy = 0.95 and $\kappa$ = 0.89). Music seems to be the most challenging domain (median bal. accuracy = 0.84, $\kappa$ = 0.69). 
The global model has a median balanced accuracy of 0.84 and a median $\kappa$ of 0.66. As expected, this is lower than the single domains models, suggesting that (some) features may not transfer well from one domain to another. 
Table \ref{tab:varimp} lists the importance of each variable in the global model and in the three single domain models, where the variable with the highest weight is ranked 1. 
The lists are relatively stable, with some marked differences, such as the importance of the gloss length and the number of partOf relations for music and fruit, respectively.




\begin{table}
  \caption{Features ranked in order of importance.}
  \label{tab:varimp}
  \centering
  \begin{tabular}{l|rrrr}
    \hline
    Feature & All & Tool & Fruit & Music \\
    \hline
depth\_in\_hierarchy   &                1     &                  1     &                  1     &                  3     \\
G.Ngrams\_2008\_max      &                2     &                  2     &                  5     &                  2     \\
gloss\_length          &                3     &                  4     &                  7     &                  1     \\
polysemy\_max        &                4     &                  3     &                  4     &                  6     \\
word\_length\_min        &                5     &                  5     &                  3     &                  4     \\
nr\_of\_partOfs           &                6     &                  8     &                  2     &                  8     \\
nr\_of\_hyponyms              &                7     &                  6     &                  6     &                  5     \\
nr\_of\_synonyms           &                8     &                  7     &                  8     &                  7     \\
nr\_of\_direct\_hypernyms          &                9     &                  9     &                  9     &                  9     \\
    \hline
  \end{tabular}
\end{table}
    
\subsubsection{Experiment 2: Prediction in a New Domain}
We explore to what extent our method can be used on new domains, for which we don't have manually labelled examples in the training set. For that purpose, we train a model on the data from two domains and test on the third domain. Table \ref{tab:exp2} shows that performance drops dramatically to $\kappa$ values between 0.27 and 0.4 depending on which domain is considered as new. This makes sense, as the distributions of feature values differ a lot over the three domains. For example, the mean gloss length is 20\% longer in the music domain. And, in our dataset, part-of relations are rare except in the Fruit domain, where 65\% of the synsets has one or more part-of relation, v.s only 10\% and 17\% in the other two.

To deal with these domain-differences, we include a normalization step. 
We apply a straightforward normalization procedure where we divide the feature value by the mean feature value within the domain.
We test performance of the classifier when either normalizing all structural features, all lexical features or the frequency feature. We find that normalization of both structural and lexical features leads to a significant performance gain. Normalization of the frequency feature seems to hurt performance (Table \ref{tab:exp2})

We conclude that basic level prediction is possible in a new domain with limited performance loss if feature values are normalized. Such a normalization step can be done automatically and does not require manual labels in the new domain. It does require, however, a decision on what constitutes a ``domain''. We have tested the hierarchies under hand tool, edible fruit and musical instruments as new domains. However, it is left for future work what would happen if one treats for example the whole hierarchy under the broad concept \verb|artefact| as a new domain, or the relatively small hierarchy under \verb|stringed instruments|. 


\begin{table*}
  \caption{Balanced accuracy and $\kappa$ of predictions made in a new domain, with or without normalization.}\label{tab:exp2}
  \label{tab:newdomain}
  \centering
  \begin{tabular}{ll|rr|rr|rr|rr}
    \hline
    \multicolumn{2}{r|}{Normalized features:} 
    & \multicolumn{2}{l|}{None} & \multicolumn{2}{l|}{Structural} & \multicolumn{2}{l|}{Lexical} & \multicolumn{2}{l}{Frequency}\\
     \hline
    New & Trained & Bal. & $\kappa$ & Bal. & $\kappa$ & Bal. & $\kappa$ & Bal. & $\kappa$\\
    domain & on  & Acc. & & Acc. & & Acc. & & Acc. & \\
    \hline
    Tools &Fruit+Music&0.69&0.40 &0.84&0.74&0.83&0.68&0.65&0.30\\
    Fruit  &Tools+Music&0.66&0.30 &0.82&0.62&0.73&0.43&0.66&0.31\\
    Music &Tools+Fruit&0.62&0.27 &0.68&0.37&0.73&0.41&0.55&0.12\\
    \hline
  \end{tabular}
\end{table*}

\subsubsection*{Experiment 3: Disagreement Among Human Annotators} 

\begin{figure}[h]
  \centering
  \includegraphics[width=0.63\linewidth]{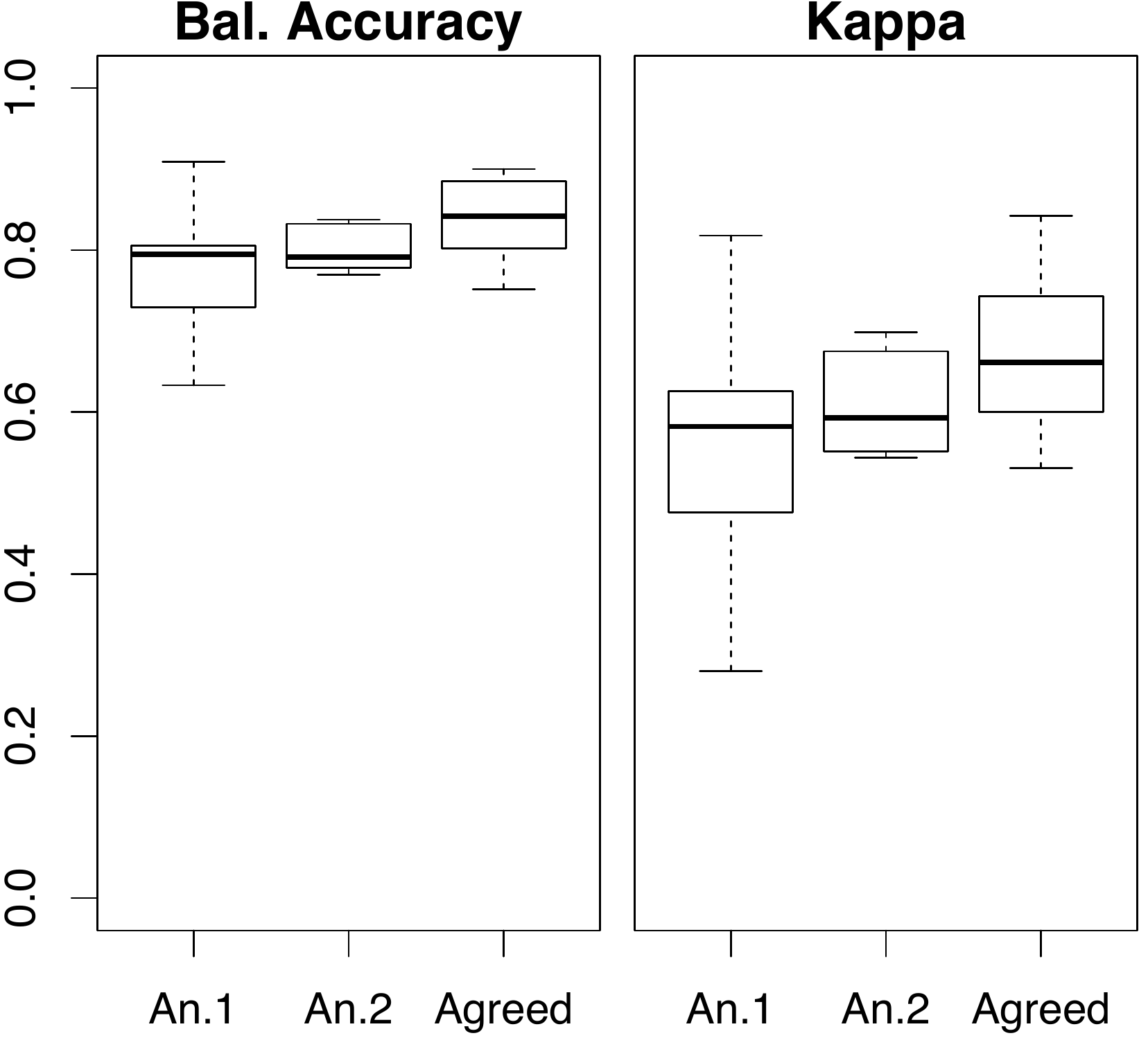}  
  \caption{
  Prediction performance on labels of annotator 1, annotator 2 or on labels that they agreed upon.} \label{fig:exp3all}
\end{figure}

The experiments above are run on the subset of the training set for which two human annotators agreed on the correct label. Here, we examine to what extent this has affected our results. We have trained and tested a global model using either the labels of annotator 1, those of annotator 2, or on the subset of synsets on which they agreed. Figure \ref{fig:exp3all} shows balanced accuracy and $\kappa$ of the three models. There are no significant differences between the two annotators ($p = 0.39$ using the Wilcoxon test on $\kappa$ values). Performance on the agreed synsets appears to be slightly higher, but the difference is not significant ($p = 0.08$ and $0.10$ when compared to annotator 1 and annotator 2, respectively).  

To gain insights into what causes the observed difference between the use of agreed synsets versus all synsets, we train a model on the agreed synsets (453 synsets), and test it on the disagreed synsets (65 synsets). When we evaluate this using the labels of annotator 1, accuracy drops to 0.62 and $\kappa$ to 0.12, an almost random classification. Evaluation on the labels of annotator 2 leads to even lower scores. 
This suggests that cases that are difficult for human annotators are also difficult for a classifier. This idea is strengthened by the results of experiment 1, in which the domain with the lowest inter-rater agreement also had the lowest classification performance, and the domain with the highest inter-rater agreement had the highest classification performance.

We hypothesize that concepts on which humans disagree are inherently difficult to classify because maybe there is no clear difference in terms of basic level effects in these cases. Concepts on which the annotators disagreed were, for example, rarely seen fruits like the sweetsop, and the sibling concepts raisin, prune and dried apricot. The concept of berry was also a cause for disagreement, where one annotator labelled berry as basic level, while the other labelled its hyponyms strawberry and blackberry as basic level. Future work will have to clarify basic level effects in these cases.

\section{Discussion and Future Work}
All three types of features proved important for basic level prediction. We believe that further improvements are possible from inclusion of additional frequency features. The Google Ngram scores that we used as frequency features gave a strong signal, and they did not need per-domain normalization. Other frequency features could include: the frequency of occurrence of words in specific corpora such as children's books or language-learning resources, and the frequency of occurrence of concepts in other conceptual hierarchies. WordNet contains also sense frequency counts, but they are available for less than 10\% of our concepts. Preliminary experiments confirmed that they are unreliable as a feature. We hypothesize that distributional models could also carry a strong signal for this task, as we expect that a vector of a basic level concept would be more similar to the vectors of its subordinates than to its superordinate or sibling concepts. Testing this hypothesis would require a inclusive set of word embeddings where polysemous words have been disambiguated, and where the bi- and trigrams found in many domains are included. Finally, we think that a link to image repositories could help. For instance, the number of examples images of a concept in ImageNet could be a signal; or the visual similarity between its example images. 

A post-hoc discussion among the annotators learned that for most concepts labelling was straightforward. A few cases were hard and annotators would have preferred to not make a choice between basic level or not. If it is true that there are concepts to which the basic level theory does not apply, it would be worthwhile to classify them a such, which would result in a classification into basic-level, not-basic-level and not-applicable. Applications will gain most from a correct classification of the straightforward cases, for which basic level effects can be expected to be strongest. In future work we intend to measure basic level effects for a larger set of concepts in a crowd-sourcing environment.

Finally, we plan to enrich knowledge graphs other than WordNet with basic level information. The structural features could then be extracted from the target knowledge graph. If the target concepts from the graph can be mapped to WordNet, lexical features could still come from WordNet.

\section{Conclusion}
We present a method to classify concepts from a conceptual hierarchy into basic level and not-basic level. We extract three types of concept features: lexical features, structural features and frequency features. We create a training and test set of manually labelled examples that includes concepts from different WordNet domains. We show that we can accurately predict the basic level within a single domain. 
The performance of a global model across multiple domains is slightly lower. Predictions in a new domain, for which there are no examples in the training set, are meaningful only after a per-domain normalization step. Concepts that are difficult to label for human annotators seem to be more challenging for the classifier as well. 


\bibliographystyle{aaai}
\bibliography{cgraph}

\begin{thebibliography}{}

\bibitem[\protect\citeauthoryear{Belohlavek and Trnecka}{2013}]{Belohlavek2013}
Belohlavek, R., and Trnecka, M.
\newblock 2013.
\newblock {Basic level in formal concept analysis: Interesting concepts and
  psychological ramifications}.
\newblock {\em IJCAI International Joint Conference on Artificial Intelligence}
   1233--1239.

\bibitem[\protect\citeauthoryear{Brown}{1958}]{brown1958shall}
Brown, R.
\newblock 1958.
\newblock How shall a thing be called?
\newblock {\em Psychological review} 65(1):14.

\bibitem[\protect\citeauthoryear{Cai \bgroup et al\mbox.\egroup
  }{2016}]{Cai2016}
Cai, Y.; Chen, W.~H.; Leung, H.~F.; Li, Q.; Xie, H.; Lau, R.~Y.; Min, H.; and
  Wang, F.~L.
\newblock 2016.
\newblock {Context-aware ontologies generation with basic level concepts from
  collaborative tags}.
\newblock {\em Neurocomputing} 208:25--38.

\bibitem[\protect\citeauthoryear{Clerkin, Cunningham, and
  Hayes}{2001}]{clerkin2001ontology}
Clerkin, P.; Cunningham, P.; and Hayes, C.
\newblock 2001.
\newblock Ontology discovery for the semantic web using hierarchical
  clustering.
\newblock {\em Semantic Web Mining} 27.

\bibitem[\protect\citeauthoryear{Corter and Gluck}{1992}]{corter1992explaining}
Corter, J.~E., and Gluck, M.~A.
\newblock 1992.
\newblock Explaining basic categories: Feature predictability and information.
\newblock {\em Psychological Bulletin} 111(2):291.

\bibitem[\protect\citeauthoryear{Golder and Huberman}{2005}]{Golder}
Golder, S.~A., and Huberman, B.
\newblock 2005.
\newblock The structure of collaborative tagging systems.
\newblock {\em Journal of Information Science} 32.

\bibitem[\protect\citeauthoryear{Green}{2006}]{Green2006}
Green, R.
\newblock 2006.
\newblock {\em {Vocabulary Alignment via Basic Level Concepts. Final Report
  2003 OCLC / ALISE Library and Information Science Research Grant Project}}.

\bibitem[\protect\citeauthoryear{Hoekstra \bgroup et al\mbox.\egroup
  }{2007}]{Hoekstra2007}
Hoekstra, R.; Breuker, J.; {Di Bello}, M.; and Boer, A.
\newblock 2007.
\newblock {The LKIF core ontology of basic legal concepts}.
\newblock In {\em CEUR Workshop Proceedings}, volume 321,  43--63.

\bibitem[\protect\citeauthoryear{Johnson and Mervis}{1997}]{Johnson1997}
Johnson, K.~E., and Mervis, C.~B.
\newblock 1997.
\newblock {Effects of Varying Levels of Expertise on the Basic Level of
  Categorization}.
\newblock {\em Journal of Experimental Psychology: General} 126(3):248--277.

\bibitem[\protect\citeauthoryear{Jones}{1983}]{jones1983identifying}
Jones, G.~V.
\newblock 1983.
\newblock Identifying basic categories.
\newblock {\em Psychological Bulletin} 94(3):423.

\bibitem[\protect\citeauthoryear{Lakoff}{2008}]{lakoff2008women}
Lakoff, G.
\newblock 2008.
\newblock {\em Women, fire, and dangerous things}.
\newblock University of Chicago press.

\bibitem[\protect\citeauthoryear{Lemaitre and Heller}{2013}]{Lemaitre2013}
Lemaitre, G., and Heller, L.~M.
\newblock 2013.
\newblock {Evidence for a basic level in a taxonomy of everyday action sounds}.
\newblock {\em Experimental Brain Research}.

\bibitem[\protect\citeauthoryear{Lin \bgroup et al\mbox.\egroup
  }{2012}]{lin2012syntactic}
Lin, Y.; Michel, J.-B.; Aiden, E.~L.; Orwant, J.; Brockman, W.; and Petrov, S.
\newblock 2012.
\newblock Syntactic annotations for the google books ngram corpus.
\newblock In {\em Proceedings of the ACL 2012 system demonstrations},
  169--174.
\newblock Association for Computational Linguistics.

\bibitem[\protect\citeauthoryear{Mark, Smith, and Tversky}{1999}]{Mark1999}
Mark, D.~M.; Smith, B.; and Tversky, B.
\newblock 1999.
\newblock {Ontology and geographic objects: An empirical study of cognitive
  categorization}.
\newblock In {\em Lecture Notes in Computer Science (including subseries
  Lecture Notes in Artificial Intelligence and Lecture Notes in
  Bioinformatics)}, volume 1661,  283--298.

\bibitem[\protect\citeauthoryear{Mathews, Xie, and
  He}{2015}]{mathews2015choosing}
Mathews, A.; Xie, L.; and He, X.
\newblock 2015.
\newblock Choosing basic-level concept names using visual and language context.
\newblock In {\em 2015 IEEE Winter Conference on Applications of Computer
  Vision},  595--602.
\newblock IEEE.

\bibitem[\protect\citeauthoryear{Miller}{1995}]{miller1995wordnet}
Miller, G.~A.
\newblock 1995.
\newblock Wordnet: a lexical database for english.
\newblock {\em Communications of the ACM} 38(11):39--41.

\bibitem[\protect\citeauthoryear{Murphy and Smith}{1982}]{Murphy1982}
Murphy, G.~L., and Smith, E.~E.
\newblock 1982.
\newblock {Basic-level superiority in picture categorization}.
\newblock {\em Journal of Verbal Learning and Verbal Behavior} 21(1):1--20.

\bibitem[\protect\citeauthoryear{Ordonez \bgroup et al\mbox.\egroup
  }{2013}]{ordonez2013large}
Ordonez, V.; Deng, J.; Choi, Y.; Berg, A.~C.; and Berg, T.~L.
\newblock 2013.
\newblock From large scale image categorization to entry-level categories.
\newblock In {\em Proceedings of the IEEE International Conference on Computer
  Vision},  2768--2775.

\bibitem[\protect\citeauthoryear{Peroni, Motta, and d'Aquin}{2008}]{PeroniMd08}
Peroni, S.; Motta, E.; and d'Aquin, M.
\newblock 2008.
\newblock Identifying key concepts in an ontology, through the integration of
  cognitive principles with statistical and topological measures.
\newblock In {\em The Semantic Web, 3rd Asian Semantic Web Conference, {ASWC}
  2008, Bangkok, Thailand, December 8-11, 2008. Proceedings},  242--256.

\bibitem[\protect\citeauthoryear{Rifkin}{1985}]{Rifkin1985}
Rifkin, A.
\newblock 1985.
\newblock {Evidence for a basic level in event taxonomies}.
\newblock {\em Memory {\&} Cognition} 13(6):538--556.

\bibitem[\protect\citeauthoryear{Rosch \bgroup et al\mbox.\egroup
  }{1976}]{Rosch1976a}
Rosch, E.; Mervis, C.~B.; Gray, W.~D.; Johnson, D.~M.; and Boyes-Braem, P.
\newblock 1976.
\newblock {Basic objects in natural categories}.
\newblock {\em Cognitive Psychology} 8(3):382--439.

\bibitem[\protect\citeauthoryear{Rosch, Simpson, and Miller}{1976}]{Rosch1976}
Rosch, E.; Simpson, C.; and Miller, R.~S.
\newblock 1976.
\newblock {Structural bases of typicality effects}.
\newblock {\em Journal of Experimental Psychology: Human Perception and
  Performance} 2(4):491--502.

\bibitem[\protect\citeauthoryear{Smith}{1967}]{Smith1967}
Smith, E.
\newblock 1967.
\newblock {Effects of Familiarity on Stimulus Recognition and Categorization}.
\newblock {\em Journal of Experimental Psychology} 74(3):324--332.

\bibitem[\protect\citeauthoryear{Tanaka and Taylor}{1991}]{Tanaka1991}
Tanaka, J.~W., and Taylor, M.
\newblock 1991.
\newblock {Object categories and expertise: Is the basic level in the eye of
  the beholder?}
\newblock {\em Cognitive Psychology} 23(3):457--482.

\bibitem[\protect\citeauthoryear{Uschold and King}{1995}]{Uschold1995}
Uschold, M., and King, M.
\newblock 1995.
\newblock {Towards a Methodology for Building Ontologies}.
\newblock {\em Methodology} 80(July):275--280.

\end{thebibliography}

\end{document}